\documentclass[letterpaper]{article} 
\usepackage{aaai2026}  
\usepackage{times}  
\usepackage{helvet}  
\usepackage{courier}  
\usepackage[hyphens]{url}  
\usepackage{graphicx} 
\usepackage{amsfonts} 
\usepackage{amsmath} 
\usepackage{times}
\usepackage{helvet}
\usepackage{courier}
\usepackage{xcolor}
\usepackage{booktabs}

\usepackage{natbib}  
\usepackage{caption} 
\usepackage{algorithm}
\usepackage{algorithmic}
\usepackage{multirow}

\usepackage{newfloat}
\usepackage{listings}
\DeclareCaptionStyle{ruled}{labelfont=normalfont,labelsep=colon,strut=off} 
\floatstyle{ruled}
\newfloat{listing}{tb}{lst}{}
\floatname{listing}{Listing}
\title{MedGR$^2$: Breaking the Data Barrier for Medical Reasoning via Generative Reward Learning}

\author{
    Weihai Zhi\textsuperscript{\rm 1,}\equalcontrib,  
    Jiayan Guo\textsuperscript{\rm 2,}\equalcontrib,    
    Shangyang Li\textsuperscript{\rm 1,3,\dag}             
}

\affiliations{
    \textsuperscript{\rm 1}Guangdong Institute of Intelligence Science and Technology, Hengqin, Zhuhai, Guangdong, China\\  
    \textsuperscript{\rm 2}School of Intelligence Science and Technology, Peking University, Beijing, China\\  
    \textsuperscript{\rm 3}Academy for Advanced Interdisciplinary Studies, Peking University, Beijing, China\\  
    
    \textsuperscript{\dag}Corresponding author. E-mail: syli@pku.edu.cn (Shangyang Li)
}

\begin{document}

\maketitle

\begin{abstract}
The application of Vision-Language Models (VLMs) in medicine is critically hampered by the scarcity of high-quality, expert-annotated data. Supervised Fine-Tuning (SFT) on existing datasets often leads to poor generalization on unseen modalities and tasks, while Reinforcement Learning (RL), a promising alternative, is stymied by the lack of reliable reward signals in this data-scarce domain. To break this impasse, we introduce Generative Reward Learning for Medical Reasoning (MedGR$^2$), a novel framework that creates a self-improving virtuous cycle. MedGR$^2$ co-develops a data generator and a reward model, enabling the automated, continuous creation of high-quality, multimodal medical data that serves as a superior training source for the post-training. Our experiments demonstrate that SFT with MedGR$^2$-produced data already surpasses baselines trained on large-scale, human-curated datasets. Crucially, when leveraging this data for RL via Group Relative Policy Optimization (GRPO), our model achieves state-of-the-art cross-modality and cross-task generalization, significantly outperforming specialized RL-based methods. Furthermore, our compact model, empowered by MedGR$^2$, achieves performance competitive with foundation models possessing over 10 times more parameters. MedGR$^2$ presents a new paradigm for data-efficient learning in high-stakes domains, transforming the problem from data scarcity to data generation and unlocking the full potential of RL for building truly generalizable medical AI.
\end{abstract}

\begin{links}
\link{Code}{https://github.com/HelmarZhi/MedGR-Square}
\end{links}

\section{Introduction}



Vision-language models (VLMs) hold great promise for transforming clinical reasoning, from multimodal diagnosis to radiology report generation. Their ability to jointly process and reason over medical images and texts offers an unprecedented opportunity to improve decision support, reduce clinician burden, and enable generalizable AI in high-stakes domains. Yet this vision is fundamentally bottlenecked by a key obstacle: the scarcity of high-quality, expert-annotated medical data~\cite{wang2025empowering, lu2025harnessing}. Compared to general-domain tasks, acquiring medical data is constrained by privacy, annotation cost~\cite{guo2023information}, and the domain-specific nature of expertise~\cite{10.1145/3737879}, which results in narrow task coverage and limited generalization across unseen modalities and tasks.

To address this challenge, the research community has explored two dominant paths, each with clear limitations. The first approach focuses on large-scale data curation and supervised fine-tuning~(SFT). Some systems~\cite{chen2024towards} attempt to clean and reformat noisy biomedical corpora such as PubMed using large vision-language models (e.g., GPT-4o), producing instruction-style datasets for downstream training. While effective at increasing data scale, these methods remain fundamentally limited by the quality of their source material and the domain alignment of the models used for cleaning. More importantly, as recent studies highlight, SFT-trained models often memorize training distributions and struggle with generalization to new modalities and tasks. This constitutes a critical gap for real-world clinical deployment.~\cite{zhao2024robust}.

The second path is reinforcement learning (RL), which offers a more powerful paradigm for learning generalizable reasoning strategies. For instance, some works\cite{lai2025med,pan2025medvlm} demonstrate that training with RL using small, expert-annotated preference datasets significantly improves cross-task and cross-modal performance over SFT baselines. However, RL in medicine faces a structural limitation: it relies on reward signals that are expensive to obtain, requiring human expert judgments for reward modeling or preference annotation. This creates a paradox: RL requires high-quality supervision to improve generalization, but such supervision is itself too costly to scale.

To break this deadlock, we propose \textbf{Generative Reward Learning for Medical Reasoning (MedGR$^2$)}, a novel self-improving framework that reframes the problem from one of data scarcity to one of data generation. Rather than curating or cleaning existing datasets, MedGR$^2$ proactively creates new multimodal training samples through a sequential pipeline: a data generator first produces candidate samples (e.g., image-question-answer triples), which are then filtered by a reward model trained to reflect expert-derived criteria such as clinical plausibility, logical consistency, and reasoning depth. The resulting high-quality samples are used to train a supervised model (via SFT), which in turn provides a stronger initialization for downstream RL. This pipeline establishes a step-wise process that transitions from raw generation to filtered supervision, and ultimately to generalizable policy learning, effectively enabling scalable and clinically robust medical reasoning. Unlike static preference annotations, our reward model evolves alongside the reasoning policy, enabling progressively refined supervision that scales without additional human effort. Our contributions are as follows:
\begin{itemize}
    \item We propose MedGR$^2$, a novel generative reward learning framework that establishes a virtuous cycle to autonomously synthesize high-quality reasoning data for medical understanding.

    \item We empirically demonstrate that our framework achieves a new state-of-the-art on the OmniMedVQA benchmark. Notably, SFT on our generated data alone already surpasses all prior fine-tuned methods, while the full MedGR$^2$ with RL pushes performance even further.

    \item We show unprecedented parameter efficiency, where our compact MedGR$^2$ model substantially outperforms foundation models over 10x its size (e.g., achieving a +19.7\% absolute gain over the 72B-parameter Qwen2-VL-72B), highlighting its immense potential for real-world clinical deployment and service.
\end{itemize}


\section{Related Work}

\subsection{Data Construction and Curation for Medical VLMs}

Constructing high-quality, large-scale datasets remains a central bottleneck in developing medical VLMs. Manual annotation is prohibitively expensive and labor-intensive, prompting a surge of semi-automated solutions. HuatuoGPT-Vision~\cite{chen2024towards} reformats PubMed-derived image-text pairs into instruction-following VQA samples using GPT-4V. LLaVA-Med~\cite{jin2023llava} synthesizes multimodal QA pairs by prompting ChatGPT with biomedical figures. PMC-VQA~\cite{wang2023pmc} leverages large language models and rule-based heuristics to mine high-quality triplets from scientific publications. Flamingo-CXR~\cite{tanno2025collaboration} demonstrates the potential of radiologist-validated datasets, yet the extensive human oversight required severely limits scalability. While these approaches offer valuable contributions, they remain constrained by source data noise, domain coverage gaps, and alignment inconsistencies. In contrast, our framework abandons static corpora altogether, employing a self-improving generative pipeline to synthesize clinically meaningful, task-aligned multimodal instruction data from scratch.

\subsection{Generative Models for Medical Data Synthesis}

To mitigate data scarcity, generative approaches such as GANs and diffusion models have been widely adopted for synthesizing realistic medical images~\cite{Yi2019GenerativeMI, chowdhury2022diffusion, 10.1145/3737879}. While effective for augmenting training sets in pixel-level tasks such as classification or segmentation, these techniques fall short in producing semantically coherent, instruction-following multimodal pairs required for medical reasoning. State-of-the-art VLMs demand data that captures clinical logic, procedural knowledge, and image-language alignment. Our framework transcends pixel-level synthesis by generating complete reasoning tasks paired with reward-guided quality filtering, enabling scalable construction of instruction-grade datasets for medical VLMs.

\subsection{Reinforcement Learning for VLM Generalization}

Although SFT has facilitated early progress in VLMs, it often encourages rote pattern learning over robust generalization~\cite{chu2025sft}. This limitation is especially detrimental in medical AI, where models must generalize to novel modalities and diagnostic scenarios. Recent studies in general domains, including One-Shot RLVR~\cite{wang2025reinforcement}, LIMO~\cite{ye2025limo}, and LIMR~\cite{li2025limr}, illustrate RL’s capacity for data-efficient learning and complex reasoning. In the medical domain, \cite{lai2025med} and MedVLM-R1 \cite{pan2025medvlm} demonstrate that RL—specifically GRPO—substantially improves generalization when guided by high-quality preferences. However, these methods rely on the availability of curated reward datasets, a costly and unsustainable requirement in clinical contexts. Our approach addresses this shortfall by integrating a generative-reward pipeline that produces instruction samples and corresponding reward signals automatically.



\section{Methodology}
\begin{figure*}[t]
    \centering
    \includegraphics[width=\linewidth]{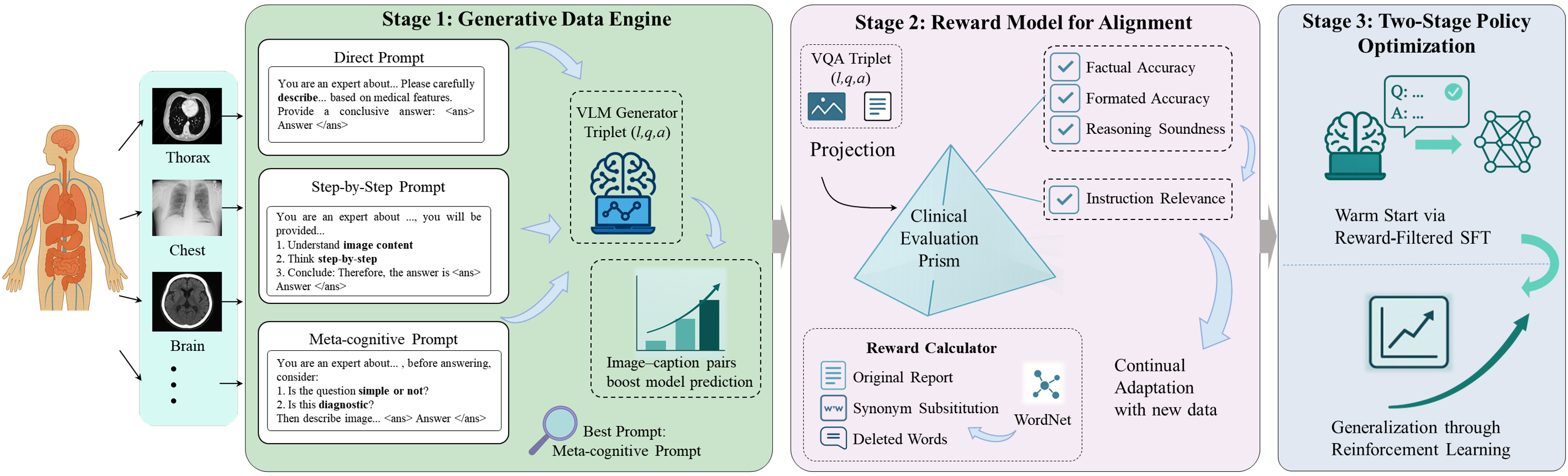} 
    \caption{\textbf{Overview of the MedGR$^2$ framework.} Our self-improving framework operates in three sequential stages. 
    Stage 1: VQA Generation. A VLM generator, guided by one of three prompt engineering strategies (Direct, Step-by-Step, Meta-cognitive), synthesizes multimodal VQA triplets $(I, q, a)$ from various medical images. Our findings indicate the Meta-cognitive prompt yields the highest quality data. 
    Stage 2: Reward Model for Alignment. A dynamic Reward Model, visualized as a "Clinical Evaluation Prism," assesses each generated triplet based on four key criteria (Factual Accuracy, Formatted Accuracy, Reasoning Soundness, Instruction Relevance). This model continually adapts using new data, providing a robust quality signal. 
    Stage 3: Two-Stage Policy Optimization. The final reasoning policy is trained in a two-stage process. First, it receives a Warm Start via Reward-Filtered SFT using high-quality data from Stage 2. Second, it is further optimized to achieve superior Generalization through RL (GRPO), creating a highly capable and robust medical reasoning model.}
    \label{fig:medgen_r1_overview} 
    \label{fig:architecture}
\end{figure*}

To address the dual challenges of \textit{data scarcity} and \textit{reasoning generalization} in medical VLMs, we propose \textbf{MedGR$^2$}, a unified, self-improving framework that couples scalable instruction generation with reinforcement-based policy optimization.Rather than a static pipeline, MedGR$^2$ functions as a dynamic ecosystem where a \textit{Multimodal Generator} $G_\theta$ and a \textit{Reward Model} $R_\phi$ co-evolve to fuel a two-stage learning process for our final Reasoning Policy $\pi_\psi$. This process begins with reward-filtered SFT for base alignment, followed by RL for superior generalization. An overview is shown in Figure~\ref{fig:architecture}.

\subsection{Multimodal Generator $G_\theta$: Engineering the Data Engine}
The generator $G_\theta$ is responsible for synthesizing diverse and clinically meaningful VQA triplets $(I, q, a)$, where $I$ is a medical image, $q$ is a diagnostic or descriptive instruction, and $a$ is a rationale-grounded answer. Unlike classical data augmentation methods, $G_\theta$ generates \textit{semantically novel} instruction-answer pairs guided by domain priors and prompt structure.

\textbf{Prompt-Driven Controllable Generation.} We implement $G_\theta$ using a powerful multimodal backbone (e.g., Gemini-2.0-Flash) and explore three prompting paradigms: (1) direct instruction, (2) step-by-step reasoning, and (3) meta-cognitive introspection. Prompt (3), which explicitly prompts the model to reflect on question types and expected answer structure, consistently yields more diverse and clinically aligned outputs. Full templates are provided in the supplementary.

\textbf{Self-Updating Generator.} The generator is periodically fine-tuned on high-reward examples filtered by $R_\phi$, allowing it to incrementally adapt to downstream reward signals and policy behaviors. This results in a self-enhancing data engine where generated samples become increasingly relevant and high-quality over iterations.

\subsection{Reward Model $R_\phi$: The Co-Evolving Expert}

The reward model $R_\phi$ serves as an automated domain expert, assigning scalar rewards $r \in [-6, 10]$ to candidate triplets $(I, q, a)$. We implement $R_\phi$ as a lightweight vision-language model followed by a regression head that predicts a reward score reflecting multiple clinically-relevant dimensions:

\begin{itemize}
\item Factual Accuracy: Is the answer clinically plausible given the image?
\item Reasoning Soundness: Is the rationale coherent and logically consistent?
\item Instruction Relevance: Does the question pertain to meaningful clinical tasks (e.g., diagnosis, description, follow-up)?
\end{itemize}

\textbf{Multi-Grade Supervision.} Unlike standard pairwise preference training, we design a four-grade reward dataset $\mathcal{D}_{\text{grade}}$ to supervise $R_\phi$ with finer control:
\begin{itemize}
\item Grade 1 (score = 10): Original reports written by radiologists.
\item Grade 2 (score = 6): Lightly perturbed versions using WordNet-based synonym replacements.
\item Grade 3 (score = 0): Incoherent answers created by random phrase deletions.
\item Grade 4 (score = -6): Irrelevant or hallucinated responses.
\end{itemize}

We minimize a mean squared error loss:
\begin{equation}
\mathcal{L}_{\text{RM}} = \mathbb{E}_{(x, r) \sim \mathcal{D}_{\text{grade}}} \left[ \left( R_\phi(x) - r \right)^2 \right]
\end{equation}

\textbf{Continual Adaptation.} Unlike static RMs trained once and deployed, $R_\phi$ in MedGR$^2$ is designed for continual adaptation. It is initialized on a small seed dataset (e.g., distilled GPT-4o preferences or expert-annotated pairs) and is periodically retrained on increasingly challenging examples generated by the improving reasoning policy $\pi_\psi$. This co-evolution enables $R_\phi$ to serve as a progressively more discerning judge, aligned with the distribution shift and complexity escalation in the generated tasks.





\subsection{Two-Stage Policy Optimization}
We train the final reasoning policy $\pi_\psi$ in two distinct but complementary stages, using the high-quality dataset $\mathcal{D}_{\text{gen}}$ filtered by $R_\phi$ (where $R_\phi(x) > \tau$).

\subsubsection{Stage 1: Warm Start via Reward-Filtered SFT}
The first stage provides a warm start by aligning $\pi_\psi$ with proven high-quality reasoning patterns. We train on $\mathcal{D}_{\text{gen}}$ with the next-token prediction objective:
\begin{equation}
\mathcal{L}_{\text{SFT}} = -\mathbb{E}_{(I, q, a) \sim \mathcal{D}_{\text{gen}}} \left[ \log \pi_\psi(a \mid I, q) \right]
\end{equation}
The primary motivation for this SFT stage is to mitigate the cold-start problem for RL. By first aligning the policy with a corpus of high-quality demonstrations, we anchor it in a region of the policy space characterized by coherent reasoning and correct formatting. This significantly stabilizes the subsequent RL exploration phase and accelerates convergence. This phase bootstraps a robust baseline model to comprehend clinically valid reasoning structure and content.

\begin{algorithm}[tb]
\caption{MedGR$^2$ Closed-Loop Training}
\label{alg:genrl}
\textbf{Input}: Seed dataset $\mathcal{D}_{\text{seed}}$, threshold $\tau$, generator $G_\theta^{(0)}$, reward model $R_\phi^{(0)}$, overlap flag \texttt{allow\_overlap}\\
\textbf{Output}: Optimized reasoning policy $\pi_\psi$
\begin{algorithmic}[1]
\STATE Initialize $\mathcal{D}_{\text{gen}} \leftarrow \mathcal{D}_{\text{seed}}$
\FOR{$t = 1, 2, \dots$}
    \STATE Generate candidate samples: $\mathcal{D}_{\text{cand}} \leftarrow G_\theta^{(t)}$
    \STATE Score all samples: $r(x) \leftarrow R_\phi^{(t)}(x),\ \forall x \in \mathcal{D}_{\text{cand}}$
    \STATE Filter high-quality samples:
        \[
        \mathcal{D}_{\text{high}} = \{x \in \mathcal{D}_{\text{cand}} \mid r(x) > \tau\}
        \]
    \IF{\texttt{allow\_overlap} is False}
        \STATE Remove overlaps: $\mathcal{D}_{\text{high}} \leftarrow \mathcal{D}_{\text{high}} \setminus \mathcal{D}_{\text{gen}}$
    \ENDIF
    \STATE Update training set: $\mathcal{D}_{\text{gen}} \leftarrow \mathcal{D}_{\text{gen}} \cup \mathcal{D}_{\text{high}}$
    \STATE \textbf{SFT:} Train $\pi_\psi$ on $\mathcal{D}_{\text{gen}}$ using loss $\mathcal{L}_{\text{SFT}}$
    \STATE \textbf{RL:} Fine-tune $\pi_\psi$ via GRPO using reward $r(a')$
    \STATE Generate preferences: $\mathcal{D}_{\text{pref}} \leftarrow \text{Sample from }\pi_\psi^{(t)}$
    \STATE Update reward model: Train $R_\phi^{(t+1)}$ on $\mathcal{D}_{\text{pref}}$
    \STATE Optionally update generator $G_\theta^{(t+1)}$ via distillation
\ENDFOR
\end{algorithmic}
\end{algorithm}

\subsubsection{Stage 2: Generalization through Reinforcement Learning}

While SFT effectively aligns the model with expert demonstrations, it often results in overly deterministic behavior, limiting the policy's robustness and adaptability. To overcome this, we introduce a RL stage that promotes exploration beyond the demonstration distribution. Rather than imitating a single ground truth, the policy $\pi_\psi$ is trained to optimize a reward signal, enabling it to uncover a broader spectrum of clinically valid reasoning trajectories. This reward-driven optimization is essential for enhancing generalization to out-of-distribution scenarios and unforeseen clinical inputs.

We employ GRPO~\cite{pan2025medvlm} to further refine $\pi_\psi$ under a composite reward that captures both reasoning quality and answer correctness. Specifically, the reward function is defined as:
\begin{equation}
r(a') = \alpha \cdot R_\phi(I, q, a') + \beta \cdot \mathbb{I}[\texttt{extract}(a') = \texttt{extract}(a)]
\end{equation}
where $R_\phi$ evaluates the semantic and clinical plausibility of the generated answer, and the indicator function $\mathbb{I}[\cdot]$ enforces alignment with the correct answer entity extracted via $\texttt{extract}(\cdot)$. This hybrid dense-plus-sparse reward structure encourages the policy to generate coherent multi-step rationales while preserving factual consistency in the final prediction.

\subsection{The Iterative MedGR$^2$ Loop}
MedGR$^2$ operates as a closed, iterative loop:
\begin{enumerate}
  \item \textbf{Generate:} The generator $G_\theta^{(t)}$ produces a new batch of candidate samples.
  \item \textbf{Filter:} The reward model $R_\phi^{(t)}$ evaluates the candidates; high scorers augment $\mathcal{D}_{\text{gen}}$.
  \item \textbf{Train Policy:} $\pi_\psi^{(t)}$ is updated via the two-stage SFT+RL process.
  \item \textbf{Evolve:} $\pi_\psi^{(t+1)}$ generates new preference pairs to refine $R_\phi^{(t+1)}$ and optionally update $G_\theta^{(t+1)}$.
\end{enumerate}
This co-evolutionary loop progressively enhances both data generation and reasoning in complex clinical domains.

\begin{table*}[t]  
\centering
\footnotesize
\renewcommand{\arraystretch}{0.98}  
\tabcolsep=5pt  
\begin{tabular}{l|cccccccc|c}
\toprule
\multicolumn{1}{c|}{\multirow{2}{*}{\textbf{Methods}}} & 
\multicolumn{8}{c|}{\textbf{Modality}} & 
\multirow{2}{*}{\textbf{Overall}} \\
\cmidrule(lr){2-9}
& \textbf{CT} & \textbf{MRI} & \textbf{X-Ray} & \textbf{US} & \textbf{Der} & \textbf{FP} & \textbf{OCT} & \textbf{Micro} & \\
\midrule
\multicolumn{10}{c}{\textbf{Zero-shot VLMs}} \\
\midrule
BLIP-2\textsuperscript{\dag} & 56.74 & 41.32 & 67.58 & 37.27 & 40.65 & 46.24 & 68.08 & 50.40 & 51.04 \\
InstructBLIP\textsuperscript{\dag} & 28.72 & 33.15 & 61.04 & 41.25 & 62.22 & 50.31 & 42.59 & 46.29 & 45.70 \\
LLaVA\textsuperscript{\dag} & 17.73 & 26.72 & 30.70 & 18.66 & 49.74 & 47.11 & 33.73 & 28.87 & 31.66 \\
LLaMA Adapter v2\textsuperscript{\dag} & 21.41 & 36.65 & 46.44 & 34.31 & 44.13 & 51.34 & 30.15 & 33.17 & 37.20 \\
MiniGPT-4\textsuperscript{\dag} & 22.81 & 27.48 & 38.30 & 25.50 & 40.52 & 33.40 & 31.36 & 36.23 & 32.54 \\
Qwen2-VL-7B & 61.46 & 40.34 & 64.27 & 51.11 & 60.02 & 63.70 & 53.34 & 53.64 & 56.40 \\
Qwen2-VL-72 & 67.97 & 69.39 & 72.11 & 51.39 & 65.31 & 72.58 & 72.76 & 67.84 & 68.05 \\
Qwen2.5-VL-7B & 60.44 & 58.44 & 73.99 & 50.46 & 62.48 & 67.66 & 67.40 & 61.87 & 60.29 \\
Qwen2.5-VL-72B & 66.18 & 63.64 & 79.81 & 69.85 & 69.75 & 71.04 & 69.22 & 69.37 & 67.71 \\
\midrule
\multicolumn{10}{c}{\textbf{Zero-shot Medical VLMs}} \\
\midrule
LLaVA-Med\textsuperscript{\dag} & 18.69 & 27.47 & 30.68 & 29.88 & 44.95 & 39.03 & 34.61 & 33.29 & 32.33 \\
RadFM\textsuperscript{\dag} & 27.56 & 24.06 & 30.95 & 16.57 & 39.21 & 36.86 & 32.80 & 27.97 & 29.50 \\
Med-Flamingo\textsuperscript{\dag} & 31.28 & 26.34 & 44.01 & 31.69 & 48.56 & 41.26 & 25.16 & 30.03 & 34.29 \\
MedVInT\textsuperscript{\dag} & 40.74 & 43.10 & 55.10 & 41.26 & 29.11 & 31.84 & 23.26 & 32.02 & 37.05 \\
\midrule
\multicolumn{10}{c}{\textbf{Fine-tuned VLMs}} \\
\midrule
Qwen2.5-VL-3B (SFT) & 66.00 & 60.81 & 69.37 & 45.11 & 62.11 & 63.95 & 65.66 & 60.83 & 61.73 \\
Qwen2.5-VL-7B (SFT) & 97.75 & 60.15 & 69.37 & 45.11 & 62.11 & 63.95 & 65.66 & 60.83 & 81.32 \\
HuatuoGPT-34B  & 60.80 & 66.50 & 83.80 & 81.70 & 74.00 & 85.50 & 90.00 & 67.40 & 76.70 \\
HealthGPT-3.8B & - & - & - & - & - & - & - & - & 68.50 \\
HealthGPT-14B & - & - & - & - & - & - & - & - & 74.40 \\
Med-R1-2B (Think) & 66.30 & 71.61 & 84.52 & 57.31 & 72.33 & 71.33 & 71.96 & 70.80 & 70.77 \\
Med-R1-3B (Nothink) & 69.89 & 72.91 & 84.52 & 43.91 & 73.62 & 80.10 & 84.18 & 71.44 & 72.57 \\
\textbf{\textit{MedGR$^2$(SFT)}} & 99.86 & 64.57& 87.52 & 47.14 & 74.34 & 73.07 & 95.62 & 71.20 & \underline{85.59}\\
\textbf{\textit{MedGR$^2$(SFT+GRPO)}}  & 99.58 & 65.00 & 92.65& 46.47& 74.55& 95.55 & 94.42&67.86 & \textbf{87.45}\\
\bottomrule
\end{tabular}
\caption{Comprehensive performance comparison of VLMs on the eight modalities of the OmniMedVQA benchmark. 
The table is segmented into three categories: zero-shot general-purpose VLMs, zero-shot medical VLMs, and fine-tuned VLMs. 
All fine-tuned models were trained on the OmniMedVQA training set for the respective modality. }
\label{tab:vlm_modalities}
\end{table*}

\section{Experiments $\And$ Results}
To evaluate the effectiveness of MedGR$^2$ in enhancing generalization and data quality for medical visual reasoning, we conduct comprehensive experiments on the OmniMedVQA benchmark.

\subsection{Research Questions}
Our experimental design is structured to answer four key research questions that progressively validate our framework's effectiveness:

\begin{itemize}
    \item \textbf{RQ1: State-of-the-Art Generalization.} Does our full MedGR$^2$ framework achieve state-of-the-art generalization performance against leading medical VLMs?

    \item \textbf{RQ2: The Synergy of Data and Reinforcement Learning.} What are the respective contributions of our high-quality generated data (via SFT) and the subsequent reinforcement learning stage (via GRPO)? 

    \item \textbf{RQ3: Beyond Accuracy: A Granular Analysis of Generalization.} How does MedGR$^2$ transfer knowledge across domains, and where does it specifically correct the errors of strong baseline models?
\end{itemize}

\paragraph{Datasets.} We use OmniMedVQA~\cite{hu2024omnimedvqa}, a large-scale, multi-specialty medical VQA benchmark. It contains over 300k triplets covering radiology, dermatology, pathology, and ophthalmology, with diverse reasoning types (e.g., diagnosis, procedural, descriptive). We use the official train/val/test splits for all experiments.

\paragraph{Task Setting.}
Our evaluation employs two distinct settings. First, for our main SOTA comparison (RQ1, RQ2), we follow the standard protocol of training models on the full, mixed-modality OmniMedVQA training set and evaluating their performance on the test set for each specific modality. Second, to specifically analyze knowledge transfer (RQ3), we adopt the stricter cross-domain protocol from Med-R1~\cite{lai2025med}, where models are trained on a single source domain (e.g., only MRI data) and evaluated on unseen target domains in a zero-shot manner.

\paragraph{Baseline Methods.}
We compare MedGR$^2$ against a comprehensive suite of baseline models, including both zero-shot and fine-tuned paradigms:

\begin{itemize}
    \item \textbf{General-Purpose VLMs (Zero-shot):} To assess the performance ceiling of large-scale vision-language models without any domain-specific tuning, we evaluate \textbf{BLIP-2}~\cite{li2023blip}, \textbf{InstructBLIP}~\cite{dai2023instructblip}, \textbf{LLaVA}~\cite{li2024llava}, \textbf{MiniGPT-4}~\cite{zhu2023minigpt}, and \textbf{LLaMA Adapter v2}~\cite{gao2023llama}, as well as strong recent models like \textbf{Qwen2-VL-7B} and \textbf{Qwen2-VL-72B}~\cite{wang2024qwen2}, and their upgraded versions \textbf{Qwen2.5-VL-7B/72B}~\cite{bai2025qwen2}.
    
    \item \textbf{Medical VLMs (Zero-shot):} We include specialized models trained on biomedical data, such as \textbf{LLaVA-Med}~\cite{li2023llava}, \textbf{RadFM}~\cite{wu2023towards}, \textbf{Med-Flamingo}~\cite{moor2023med}, and \textbf{MedVInT}~\cite{zhang2024development}, to benchmark domain-specific zero-shot capabilities.

    \item \textbf{Fine-tuned VLMs:} This group contains all models that are fine-tuned on the OmniMedVQA training split.
    \begin{itemize}
        \item \textbf{SFT Baselines:} We fine-tune strong general-purpose models such as \textbf{Qwen2.5-VL-3B} and \textbf{Qwen2.5-VL-7B} using standard supervised instruction tuning. We also consider curated data generation pipelines like \textbf{HuatuoGPT-Vision}~\cite{chen2024towards} and \textbf{HealthGPT-Vision}\cite{lin2025healthgpt} when available.
        
        \item \textbf{RL Baselines:} We include \textbf{Med-R1}~\cite{lai2025med} variants (2B and 3B, with and without reasoning modules) as state-of-the-art reward-learning based VLMs, and compare directly against our own model variants: \textbf{MedGR$^2$ (SFT)} and \textbf{MedGR$^2$ (SFT+GRPO)}.
    \end{itemize}
\end{itemize}

\paragraph{Evaluation Metrics.} We report top-1 Accuracy as the primary metric for all VQA tasks. To provide a more nuanced view on class-imbalanced diagnostic tasks, we also report Balanced Accuracy and AUROC where applicable.

\paragraph{Implementation Details.} Our reasoning policy $\pi_\psi$ is based on the Qwen2.5-VL-7B architecture. The generator $G_\theta$ utilizes the Gemini-2.0-Flash API with the meta-cognitive prompt identified as most effective. The MedGR$^2$ framework was run for 3 iterative cycles, generating approximately 50k VQA samples per cycle. The reward model $R_\phi$ is initialized from a pre-trained 3B VLM and fine-tuned on an initial seed set of 2k preference pairs distilled from GPT-4o. For RL, we use the GRPO implementation from Med-R1 with reward coefficients $\alpha=0.8$ and $\beta=0.2$. All models were trained on a cluster of 8x NVIDIA L40 GPUs. Further hyperparameters are detailed in the Appendix.

\subsection{RQ1: State-of-the-Art Performance Across Medical Modalities}

To answer our first research question, we conduct a large-scale comparison of MedGR$^2$ against a diverse set of leading VLMs on the OmniMedVQA benchmark. Table~\ref{tab:vlm_modalities} reports the detailed accuracy across eight clinically diverse imaging modalities.

\textbf{Overall State-of-the-Art Performance.}
Our full model MedGR$^2$ (SFT+GRPO) achieves the highest overall accuracy of 87.45\%, outperforming all baselines by a significant margin. Compared to the 72B-parameter foundation model Qwen2.5-VL-72B (67.71\%), our 7B MedGR$^2$ yields a +19.74\% absolute gain, demonstrating that a well-designed generative and reward-guided framework can surpass models more than 10$\times$ larger in scale.

\textbf{Cross-Specialty Generalization.}
MedGR$^2$ excels across a wide range of modalities with top-tier performance in highly heterogeneous domains such as X-Ray (92.65\%), Fundus Photography (95.55\%), and OCT (94.42\%). This consistent dominance reflects our method’s ability to generate semantically rich, clinically grounded supervision signals that generalize beyond narrow modality-specific distributions.

\textbf{The Power of Synthetic Data.}
Even without reinforcement learning, MedGR$^2$ (SFT)—trained purely on generated data—achieves 85.59\% accuracy, outperforming all fine-tuned baselines, including HuatuoGPT-34B (76.70\%) and the RL-based Med-R1-3B (72.57\%). This validates that our generator-reward pipeline produces training samples of higher fidelity and diversity than human-curated datasets.

\textbf{RL Amplifies Generalization.}
Applying GRPO on top of the SFT model yields a further boost to 87.45\%, highlighting the synergistic effect of reinforcement learning when trained in a reward-informed data ecosystem. Unlike static preference tuning, our reward model evolves alongside the reasoning policy, enabling progressively refined supervision without additional human effort.

These results challenge the "scale-is-all-you-need" hypothesis in medical AI. MedGR$^2$ demonstrates that generalization can be achieved more efficiently through intelligent data generation and optimization rather than brute-force scaling, thereby paving the way for practical, high-performing VLMs deployable in real-world healthcare settings.

\subsection{RQ2: The Synergy of High-Quality Data and Reinforcement Learning}
\begin{figure}[htbp]
    \centering
    \includegraphics[width=\linewidth]{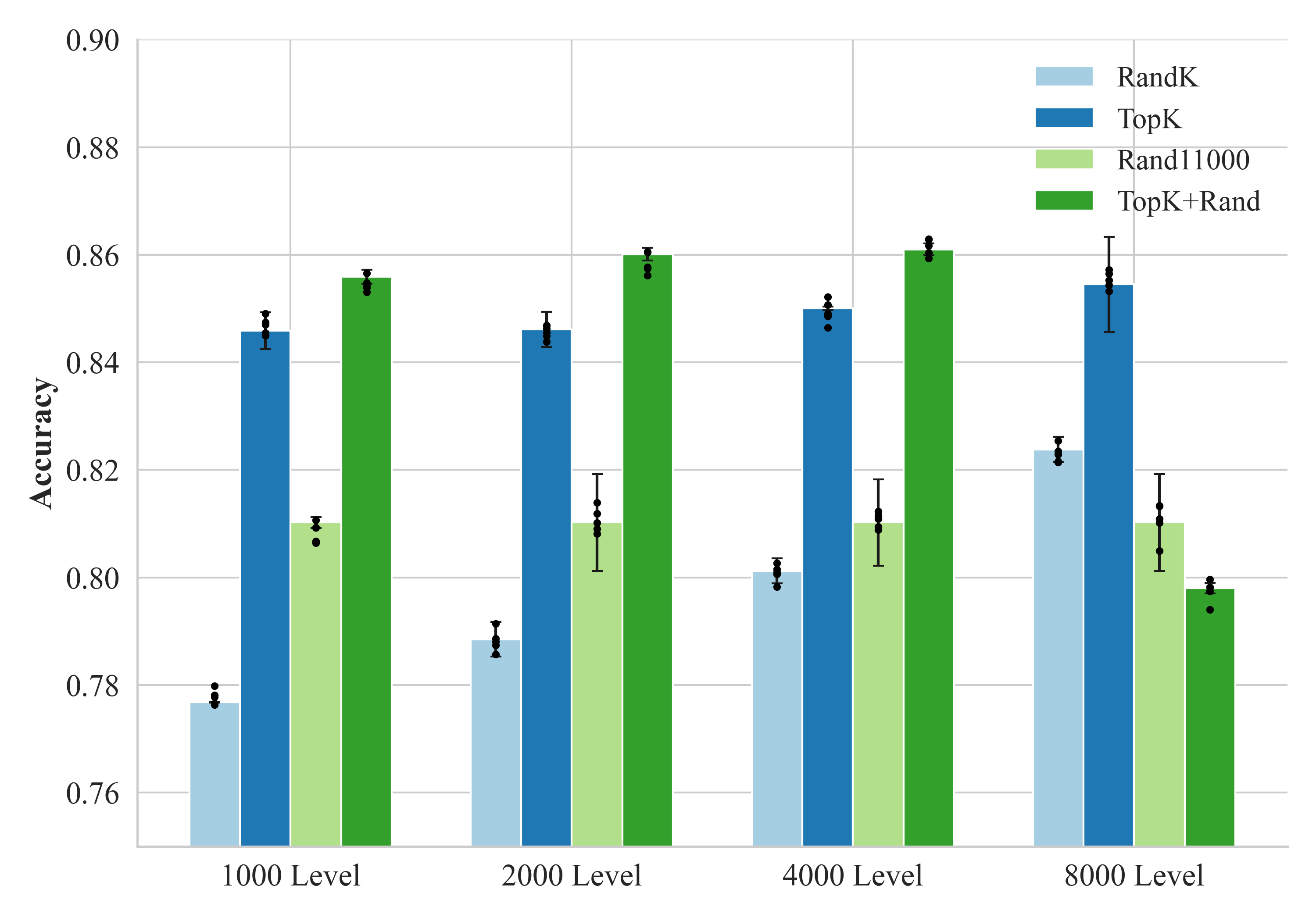} 
    \caption{The synergy of reward-filtered data and Reinforcement Learning across varying data scales. We compare four training strategies: `RandK` (SFT on K randomly sampled generated data); `TopK` (SFT on the top K samples selected by our reward model; `Rand11000` (a baseline SFT on 11,000 unfiltered samples); and `TopK+Rand` (MedGR$^2$ with GRPO applied on the `TopK` data). 
    }
    \label{fig:data_vs_rl_synergy} 
\end{figure}

\begin{table}[htbp]
\centering
\resizebox{\columnwidth}{!}{
\begin{tabular}{l|ccc|ccc}
\toprule
\textbf{Models} & \multicolumn{3}{c|}{\textbf{OmniMed}} & \multicolumn{3}{c}{\textbf{MMMU}} \\
 & Acc & F1 & AUC & Acc & F1 & AUC \\
\midrule
Qwen2.5-VL & 73.63 & 68.10 & 69.55 & 69.42 & 67.35 & 71.88 \\
+ SFT (raw GT answer) & 81.32 & 80.21 & 76.74 & 73.18 & 71.02 & 75.43 \\
+ SFT (rewarded pairs) & 85.56 & 84.19 & 83.92 & 76.83 & 74.95 & 78.30 \\
+ GRPO & \textbf{87.45} & \textbf{-} & \textbf{-} & \textbf{78.66} & \textbf{76.51} & \textbf{79.88} \\
\bottomrule
\end{tabular}
}
\caption{Ablation study on OmniMed and MMMU datasets. We evaluate models using Accuracy, F1, and AUC. Reward-based SFT and GRPO provide consistent gains over raw SFT baselines.}
\label{tab:ablation}
\end{table}

To better understand the components behind MedGR$^2$’s superior performance, we investigate RQ2 by analyzing the respective contributions of our reward-filtered data and the RL stage. Our goal is to characterize how each component contributes to performance gains across both varying data scales and datasets.

\paragraph{Reward-Filtered Data Enables Extreme Data Efficiency.}
Figure~\ref{fig:data_vs_rl_synergy} evaluates the performance of SFT models trained on different subsets of generated data. We compare training with: (i) randomly selected K samples (`RandK`), (ii) top-K high-reward samples selected by our reward model (`TopK`), and (iii) a large baseline set of 11,000 raw, unfiltered samples (`Rand11000`). Across all sample sizes (1k–8k), TopK consistently outperforms RandK by a large margin. Notably, `TopK` with only 2,000 samples already achieves 84.6\% accuracy—surpassing the large-scale `Rand11000` baseline. This result highlights the exceptional quality and data efficiency of our generative reward filtering process, which effectively selects high-value training samples without human involvement.

\paragraph{Reinforcement Learning Unlocks Robust Reasoning.}
While high-quality data provides a strong training foundation, our full framework leverages RL via GRPO to further enhance model generalization. Table~\ref{tab:ablation} presents ablation results across two benchmarks: OmniMedVQA and MMMU.

We observe a clear, step-wise gain: supervised fine-tuning on raw GT pairs yields an accuracy of 81.32\% on OmniMedVQA. Using reward-filtered QA pairs improves this to 85.56\%. Applying GRPO on top of the SFT model brings the final accuracy to 87.45\%, confirming the strong synergy between reward-guided data and reinforcement learning. On MMMU, a similar trend is observed:the full pipeline achieves 78.66\% accuracy, outperforming the raw SFT baseline by 5.48\%. These improvements validate our hypothesis that reward-driven generation and exploration-based RL form a complementary, self-reinforcing training loop, boosting the overall performance.

\paragraph{When Does RL Matter Most?}
Figure~\ref{fig:data_vs_rl_synergy} further reveals an interesting trend: RL provides the most benefit when the amount of reward-filtered data is small to moderate (e.g., 1k–4k). At 8k, however, SFT-only models slightly outperform their RL-augmented counterparts. This suggests a potential saturation point where sufficient high-quality data enables effective imitation learning, and the marginal benefits of further RL diminish. Thus, we recommend applying GRPO in scenarios with limited, high-quality supervision, which is common in many specialized medical domains.

\subsection{RQ3: A Granular Analysis of Generalization and Its Origins}

\begin{figure}[htbp]
    \centering
    \includegraphics[width=0.55\linewidth]{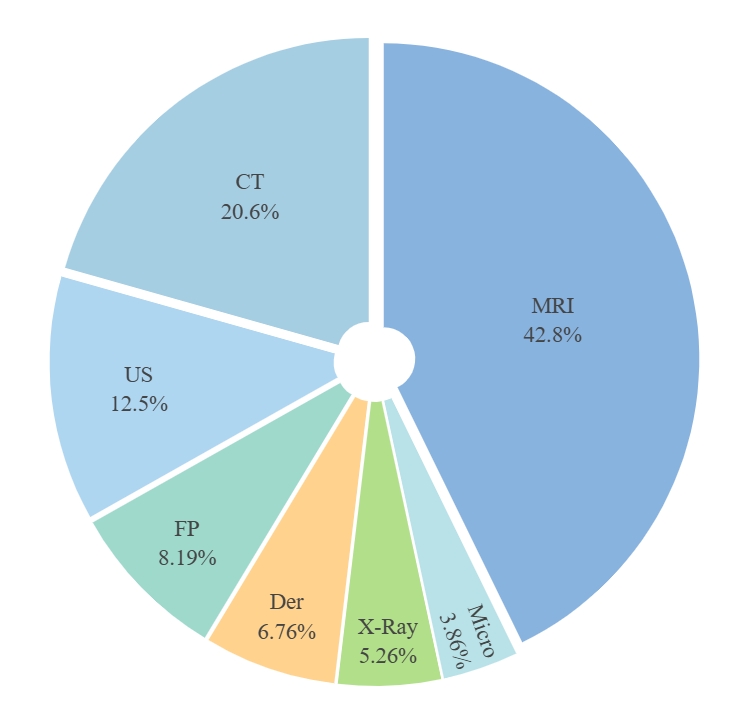}
    \caption{ Distribution of modalities in all training dataset. }
    \label{fig:modality-pie}
\end{figure}

To further understand the sources of MedGR$^2$’s effectiveness, we investigate how it generalizes across tasks and modalities. Our analysis focuses on two key factors: (1) transferability from imbalanced training data and (2) fine-grained error correction dynamics.
Figure~\ref{fig:modality-pie} shows the training data distribution: modalities like MRI (42.8\%) and CT (20.6\%) dominate, while Microscopy (3.86\%) is underrepresented. Despite this imbalance, MedGR$^2$ exhibits non-trivial transfer patterns.

\begin{figure}[htbp]
    \centering
    \begin{minipage}[t]{0.49\linewidth}
        \centering
        \includegraphics[width=\linewidth]{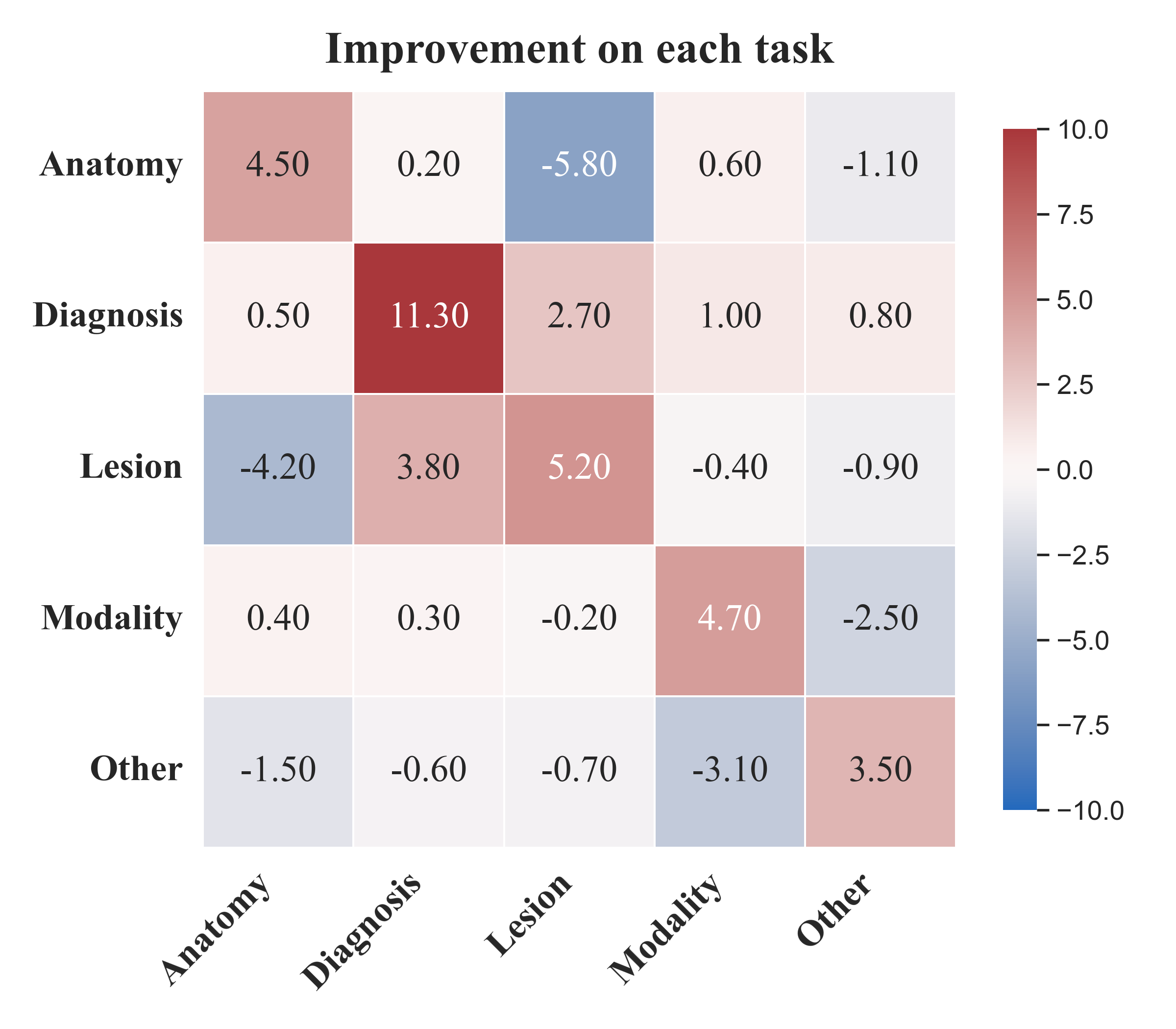}
        \caption*{(a) Cross-task transfer performance.}
    \end{minipage}%
    \hfill
    \begin{minipage}[t]{0.49\linewidth}
        \centering
        \includegraphics[width=\linewidth]{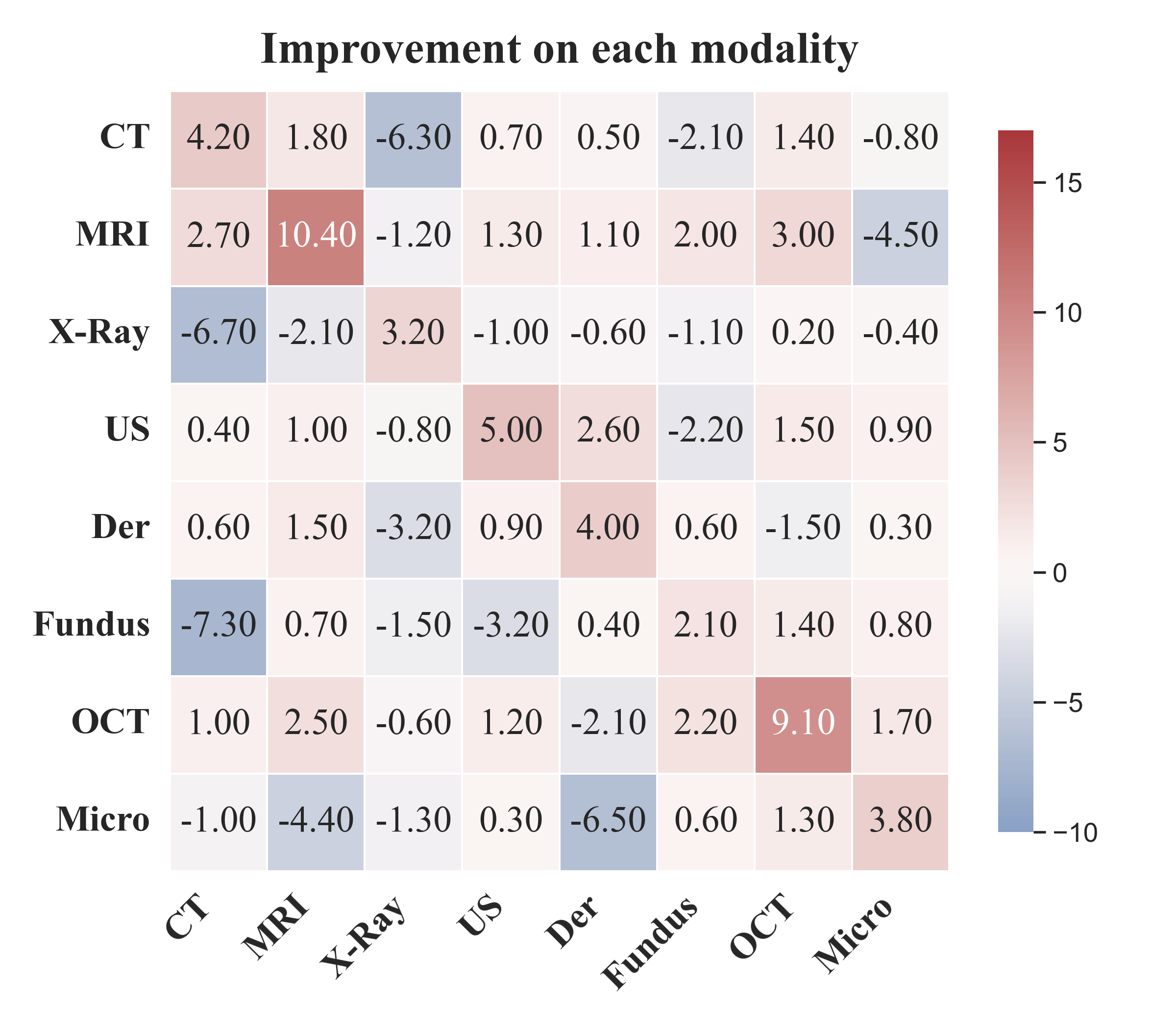}
        \caption*{(b) Cross-modality transfer performance.}
    \end{minipage}
    \caption{Cross-task and cross-modality transferability. Warmer colors indicate positive transfer, while cooler tones show negative effects.}
    \label{fig:cross_transfer_heatmaps}
\end{figure}

\subsubsection{Transferability Across Tasks and Modalities}

Figure~\ref{fig:modality-pie} shows the training data distribution: modalities like MRI (42.8\%) and CT (20.6\%) dominate, while Microscopy (3.86\%) is underrepresented. Despite this imbalance, MedGR$^2$ exhibits non-trivial transfer patterns.

The task transfer heatmap (Figure~\ref{fig:cross_transfer_heatmaps}a) reveals that training on Diagnosis leads to broad positive transfer across tasks, suggesting it captures foundational reasoning skills. Meanwhile, the modality transfer heatmap (Figure~\ref{fig:cross_transfer_heatmaps}b) shows that frequent modalities like MRI enable strong transfer to others like CT and Microscopy, indicating that MedGR$^2$ learns generalizable features rather than memorizing frequency.

\subsubsection{Error Correction Breakdown}

\begin{figure}[t]
    \centering
    \includegraphics[width=\columnwidth]{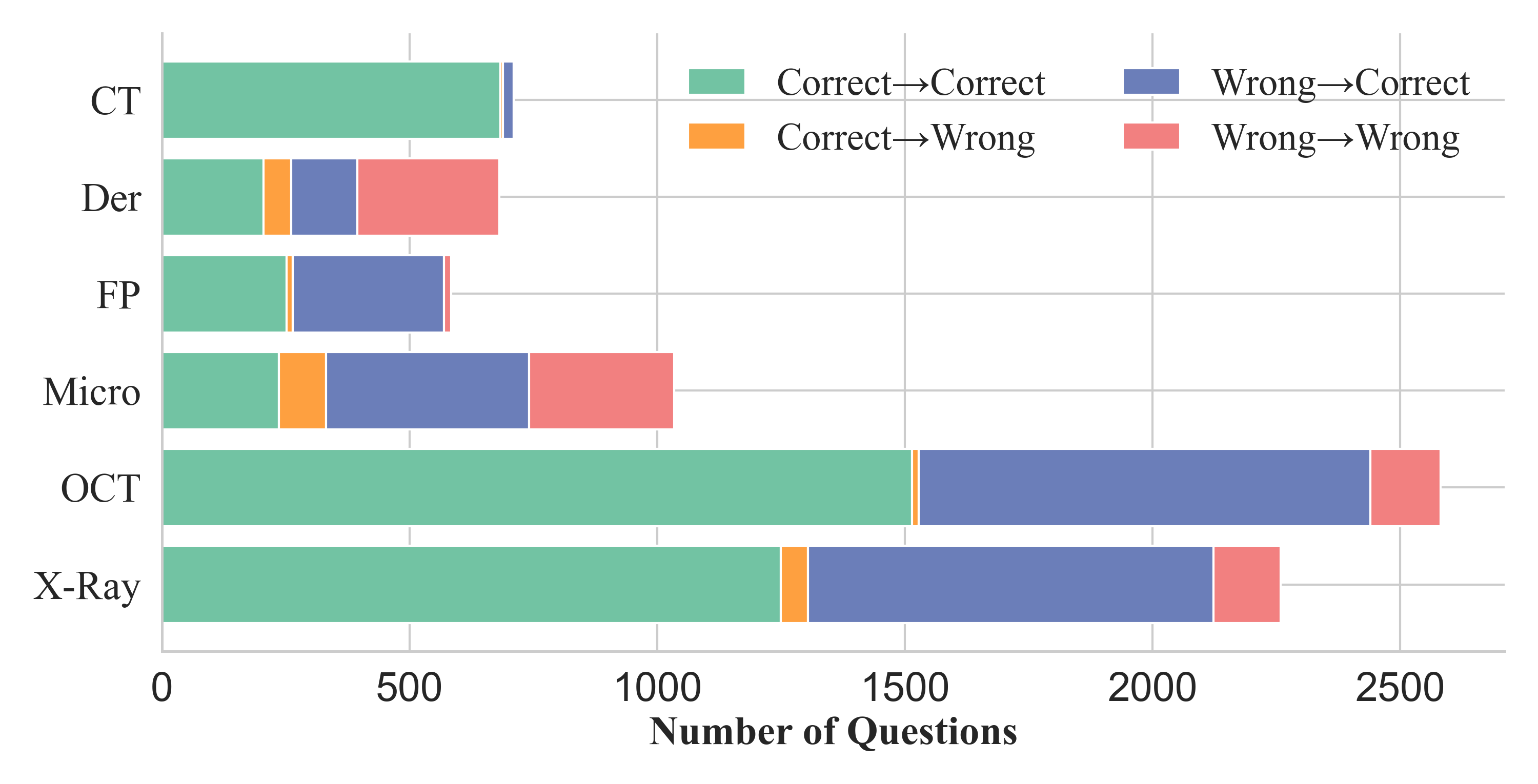} 
    \caption{Head-to-Head Error Transition Analysis. 
    Each bar dissects the entire question set for a modality into four transition flows based on the outcomes of the two models. 
    The four flows are: Correct$\rightarrow$Correct (green), where both models succeed; Wrong$\rightarrow$Correct (blue), where MedGR$^2$ corrects the baseline's error; Correct$\rightarrow$Wrong (orange), where our model makes a new error on a question the baseline answered correctly; and Wrong$\rightarrow$Wrong (red), where both models fail.}
    \label{fig:error_transition_analysis}
\end{figure}

To analyze how these capabilities translate into performance gains, we perform a head-to-head comparison with a strong SFT baseline. Figure~\ref{fig:error_transition_analysis} visualizes the transitions: correct-to-correct (green), wrong-to-correct (blue), correct-to-wrong (orange), and wrong-to-wrong (red).

MedGR$^2$ consistently yields more corrected errors (blue) than new ones (orange) across all modalities. Particularly in high-resource domains like X-Ray and OCT, the number of baseline errors fixed by MedGR$^2$ is substantial. Even in low-resource settings (e.g., Dermatology), it shows meaningful improvements. These results confirm that MedGR$^2$ not only generalizes well but also systematically reduces error rates, showcasing its robust and transferable reasoning ability.


\section{Conclusion}
In this work, we introduced MedGR$^2$, a novel framework addressing data scarcity and generalization challenges in medical VLM via a self-improving virtuous cycle of data generation and reward modeling. Our experiments show MedGR$^2$ achieves new state-of-the-art performance on the OmniMedVQA benchmark, outperforming strong SFT and RL baselines, and even substantially larger foundation models, demonstrating remarkable parameter efficiency. By transforming the core problem from data curation to intelligent, automated data generation, MedGR$^2$ paves the way for more robust, generalizable, and scalable AI for high-stakes clinical reasoning.

\section{Acknowledgments}
This research was Supported by Guangdong S\&T Program under Grant 2023B0303040002, and the Young Scientists Fund of the National Natural Science Foundation of China
(C-Class, No. 32500997). The authors would like to express their sincere gratitude to Prof. Lirong Zheng and Prof. Yuxiang Huan for their constructive discussions and valuable feedback. The authors also thank all members of the lab for their continuous support.

\bibliography{aaai2026}

\end{document}